\documentclass[journal]{IEEEtran}
\usepackage{amssymb}
\usepackage{multirow}
\usepackage{booktabs} 
\usepackage{graphicx}
\usepackage{url} 
\usepackage{bbding}
\usepackage{pifont}
\usepackage{fontawesome}
\usepackage{wasysym}
\usepackage{utfsym}

\ifCLASSINFOpdf

\else

\fi

\begin{document}

\title{Fine-Grained Building Function Recognition from Street-View Images via Geometry-Aware Semi-Supervised Learning}

%

% \author{Michael~Shell,~\IEEEmembership{Member,~IEEE,}
%         John~Doe,~\IEEEmembership{Fellow,~OSA,}
%         and~Jane~Doe,~\IEEEmembership{Life~Fellow,~IEEE}% <-this % stops a space
% \thanks{M. Shell is with the Department
% of Electrical and Computer Engineering, Georgia Institute of Technology, Atlanta,
% GA, 30332 USA e-mail: (see http://www.michaelshell.org/contact.html).}% <-this % stops a space
% \thanks{J. Doe and J. Doe are with Anonymous University.}% <-this % stops a space
% \thanks{Manuscript received April 19, 2005; revised September 17, 2014.}}

\author{Weijia~Li,
        Jinhua~Yu,
        Dairong~Chen,
        Yi~Lin,
        Runmin~Dong,
        Xiang~Zhang,
        Conghui~He
        and~Haohuan~Fu% <-this % stops a space
\thanks{This work was supported in part by the National Natural Science Foundation of China under Grant 42201358. (Weijia Li and Jinhua Yu
contributed equally to this work.) (Corresponding authors: Runmin Dong; Haohuan Fu.)}.
\IEEEcompsocitemizethanks{\IEEEcompsocthanksitem Weijia Li, Jinhua Yu, Dairong Chen, Yi Lin, and Xiang Zhang are with the School of Geospatial Engineering and Science, Sun Yat-sen University, Zhuhai 519082, China.
E-mail: \{liweij29,zhangx795\}@mail.sysu.edu.cn, \{yujh56,chendr7,liny377\}@mail2.sysu.edu.cn.
\IEEEcompsocthanksitem Runmin Dong is with the Ministry of Education Key Laboratory for Earth System Modeling, Department of Earth System Science, Tsinghua University, Beijing 100084, China.
E-mail: drm@mail.tsinghua.edu.cn.
\IEEEcompsocthanksitem Conghui He is with Shanghai Artificial Intelligence Laboratory, Shanghai 200030, China, and also with SenseTime Research, Shenzhen 518038, China.
E-mail: heconghui@pjlab.org.cn.
\IEEEcompsocthanksitem Haohuan Fu is with Tsinghua Shenzhen International Graduate School, Tsinghua University, Shenzhen 518055, China, and also with the Ministry of Education Key Laboratory for Earth System Modeling, and the Department of Earth System Science, Tsinghua University, Beijing 100084, China.
E-mail: haohuan@tsinghua.edu.cn.
}
}

% \markboth{IEEE TRANSACTIONS ON GEOSCIENCE AND REMOTE SENSING, Vol. 1, September, 2024}%
% {LI, \MakeLowercase{\textit{(et al.)}}: Fine-Grained Building Function Recognition from Street-View Images via Geometry-Aware Semi-Supervised Learning}



\maketitle

\begin{abstract}
The diversity of building functions is vital for urban planning and optimizing infrastructure and services. Street-view images offer rich exterior details, aiding in function recognition. However, detailed street-view building function annotations are limited and challenging to obtain. In this work, we propose a geometry-aware semi-supervised framework for fine-grained building function recognition, utilizing geometric relationships among multi-source data to enhance pseudo-label accuracy in semi-supervised learning, broadening its applicability to various building function categorization systems. Firstly, we design an online semi-supervised pre-training stage, which facilitates the precise acquisition of building facade location information in street-view images. In the second stage, we propose a geometry-aware coarse annotation generation module. This module effectively combines GIS data and street-view data based on the geometric relationships, improving the accuracy of pseudo annotations. In the third stage, we combine the newly generated coarse annotations with the existing labeled dataset to achieve fine-grained functional recognition of buildings across multiple cities at a large scale. Extensive experiments demonstrate that our proposed framework exhibits superior performance in fine-grained functional recognition of buildings. Within the same categorization system, it achieves improvements of 7.6\% and 4.8\% compared to fully-supervised methods and state-of-the-art semi-supervised methods, respectively. Additionally, our method also performs well in cross-city scenarios, i.e., extending the model trained on OmniCity (New York) to new cities (i.e., Los Angeles and Boston) with different building function categorization systems. This study offers a new solution for large-scale multi-city applications with minimal annotation requirements, facilitating more efficient data updates and resource allocation in urban management.
\end{abstract}

\begin{IEEEkeywords}
Building function, Semi-supervised learning, Object detection, Street-view images, GIS data
\end{IEEEkeywords}

\IEEEpeerreviewmaketitle

\section{Introduction}

\IEEEPARstart{B}{uildings} are a fundamental component of urban infrastructure, with their functions directly linked to the economic and cultural development of cities \cite{srivastava2019understanding}, attracting considerable scholarly attention \cite{gonzalez2012automated,zhou2023building,li2023joint,tong2022relationships}. As urban modernization progresses, the demand for urban planning and management continues to grow, making the efficient and intelligent understanding of building functions critical for comprehending urban structures and assisting in planning and management \cite{jiao2021hidden}. However, the fine-grained functional information of buildings in some areas currently suffers from issues such as poor timeliness and missing information. This is mainly due to the fact that the majority of gaining the function data still rely on costly manual collection, which is not conducive to efficient urban understanding and management. Therefore, researching rapid and intelligent methods for large-scale fine-grained urban building function recognition is of significant importance.

Street-view images have become a very popular ground-level data source in recent years, accurately mapping the physical spaces of urban streets and demonstrating substantial potential for building function recognition\cite{workman2022revisiting,fang2022spatial,ye2024sg,ye2024cross,sun2023cross}. There have been abundant annotations for non-building objects in street-view images proposed by current studies, such as cars, pedestrians, and road signs \cite{geiger2012we,caesar2020nuscenes,cordts2016cityscapes,wang2022detecting}. However, annotations for buildings, especially fine-grained functional annotations, are relatively lacking. This is because buildings are large in size, and deformations are more pronounced in street-view images, especially in panoramic views, which increases the difficulty of building annotation. In addition, most existing street-view annotation data are manually labeled, which is costly and inefficient \cite{zhou2020holicity,yang2019pass,zhao2021bounding}. For data-dependent supervised models, this is not conducive to achieving high-precision and large-scale recognition of urban building functions. In contrast, satellite images and GIS data from the top-down perspective are relatively abundant in annotations, such as building footprints and attribute information \cite{chen2018social,chen2023large}. However, due to differences in data collection perspectives and potential positioning errors, how to effectively combine these two types of data to automatically generate annotations in the street-view perspective is a challenging problem that warrants further research.

Fully-supervised deep learning methods are the mainstream approach for fine-grained building function recognition \cite{li2023omnicity,kang2018building}. However, the existing labeled information does not meet the high demands for data quantity and quality required by fully supervised methods. Semi-supervised learning combines the ideas of both supervised and unsupervised learning, effectively reducing the model's dependency on labeled data. The most common semi-supervised learning framework is based on pseudo-label generation, which, through the assumption between predicted samples and learning targets, requires only a small amount of labeled data. This approach extends the knowledge learned from the known domain to the target domain in the form of pseudo-labels \cite{xu2021end,wang2023semi}. Scholars have already applied this method to urban tasks with promising results \cite{guo2021deep,li2022semi,yu2022pixel,hong2020x}. However, for more complex task scenarios such as fine-grained building function recognition, semi-supervised methods relying on pseudo-label generation still face challenges such as low pseudo-label accuracy and limited applicability to a single classification system.

To enhance the accuracy of pseudo labels and effectively utilize multi-source urban data, we propose a geometry-aware semi-supervised framework for fine-grained building function recognition, as shown in Figure \ref{fig:pipeline}. This framework effectively utilizes GIS data and street-view images, enhancing the accuracy of pseudo-label generation in semi-supervised framework, and reduces the model's dependence on annotated samples. Specifically, it mainly includes three stages, building facade detection, building function annotation generation, and fine-grained building function recognition. In the first stage, we employ a semi-supervised framework for pre-training, which requires only a small amount of labeled data to obtain a pre-trained model capable of detecting building facades over a wide area. In the second stage, we introduce a coarse annotation generation module that leverages geometric angular relationships to align building function information from GIS data with building facade detected by the pre-trained model. In the third stage, we reorganize the existing labeled data and the coarse annotations obtained from the second stage, feeding them into a single-stage object detection network, which accomplished fine-grained building function recognition. Our framework outperforms fully-supervised methods by 7.9\% and achieves the best performance among state-of-the-art semi-supervised methods in New York City, using only 10\% of the labeled data. Furthermore, unlike previous semi-supervised methods, the multi-stage training strategy enables our method to accomplish pixel-level fine-grained building function recognition in cross-categorization systems. %(i.e., New York - Los Angeles and New York - Boston). 

Our main contributions are summarized as follows:

\begin{figure*}[ht]
    \centering
    \includegraphics[width=\linewidth]{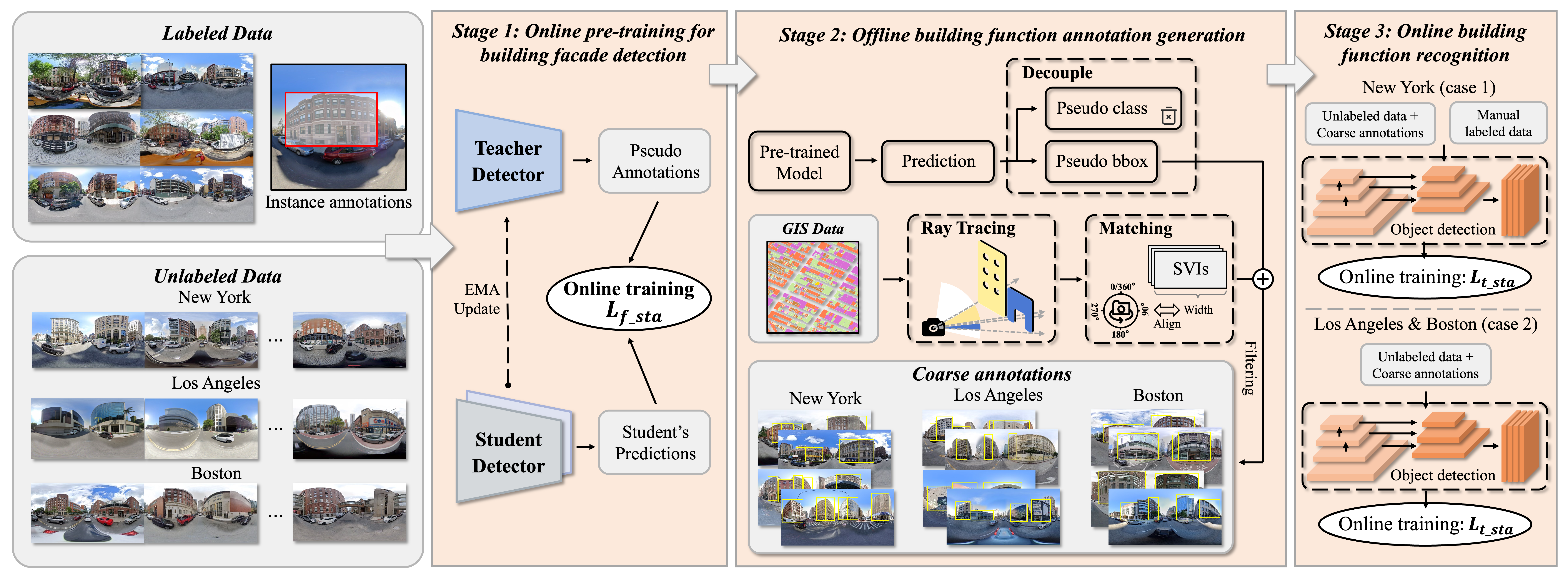}
    \caption{An overview of the proposed geometry-aware semi-supervised framework, which include: (a) An online semi-supervised object detection framework for building facade detection, (b) a brand new module for coarse annotation generation based on GIS data and street-view images, (c) One-stage-based building function recognition.}
    \label{fig:pipeline}
\end{figure*}

\begin{itemize}
\item We propose a geometry-aware semi-supervised method for fine-grained building function recognition. This method enables large-scale function recognition with a small number of samples.
\item We propose a coarse annotation generation module that effectively utilizes multi-source urban data. By leveraging the geometric relationship between different perspectives of GIS data and street-view data, this module enhances the accuracy of pseudo-labels.
\item We construct datasets for new cities, i.e., Los Angeles and Boston, based on the CityAnnotator proposed in our previous work \cite{li2023omnicity}. Experimental results show that our method exhibits strong robustness in cross-categorization system tasks, enabling the acquisition of large-scale, multi-city fine-grained building function data.
\end{itemize}

\section{Related work}
\subsection{Fine-grained building attributes recognition based on street-view images}

Many scholars have combined street-view data with deep learning methods to conduct urban fine-grained classification tasks. These classification tasks can be divided into two categories based on different attributes of buildings: geometric-physical attributes and socio-economic functions of buildings.
For the first aspect, the geometric-physical attributes of buildings include height, shape, and density, with the estimation of building height being a key element in the study of urban three-dimensional morphology \cite{pang20223d,zhao2022scalable}. Some studies \cite{yan2022estimation, xu2023building} used fully convolutional networks (FCNs) to identify building instances from street-view images and then mapped them to real-world buildings based on geometric relations to achieve accurate height estimation. Other studies \cite{szczesniak2022method,hu2022semi,kong2020enhanced} analyzed the facade structures of buildings based on street-view images, thoroughly describing the shapes of buildings' windows and doors. 
Zhou et al. \cite{zhou2020holicity} proposed a street-view dataset, generated various annotations for the overall three-dimensional structure of buildings based on CAD, and provided benchmark experiments for multiple tasks.

The socio-economic function of buildings provide a deeper understanding of urban and can be further detailed into two types. The first type of socio-economic function refers to an assessment of the individual functionality of a single building. For instance, some researchers used street-view data to sort urban buildings into categories like Residential, Commercial, and Industrial \cite{hoffmann2019model,ramalingam2023automatizing}. Then, Kang et al. \cite{kang2018building} used several street-view images from different angles of the same building for a more precise land use classification. Meanwhile, some studies \cite{sun2021automatic,ogawa2023deep} applied deep learning methods to estimate building ages, there were also some studies used street-view images to automatically predict buildings' energy use \cite{rosenfelder2021predicting}, materials \cite{wang2021automatic}, and earthquake safety \cite{pelizari2021automated}. The second type of socio-economic function considers the building entities in conjunction with the city environment. For instance, Hu et al. \cite{hu2020classification} utilized street-view images and deep learning methods to propose an urban canyon classification framework, considering building heights and road widths, while Zhao et al. \cite{zhao2023quantitative} estimated property prices by taking into account buildings, human factors, and environmental factors.

In summary, the rich and detailed building facade features in street-view images can effectively improve the accuracy of fine-grained building function recognition, especially for socio-economic functions. However, fine-grained functional annotation data for buildings from the street-view perspective is extremely scarce. This is because the current street-view annotation process still relies on costly manual annotation. The scarcity of fine-grained building functional annotation data directly leads to the fact that fully supervised learning methods, which heavily rely on annotated data, have difficulties in achieving high-precision large-scale building function recognition.

\subsection{Semi-supervised learning}
Self-training with pseudo-labeling and its variations has achieved significant success in the semi-supervised learning field \cite{sohn2020simple,tarvainen2017mean,xu2021end,liu2021unbiased,wang2023consistent}. Semi-supervised frameworks based on pseudo-labeling are generally implemented through the use of a teacher-student architecture. The teacher-student architecture defines two models, namely the teacher model and the student model. The teacher model is generated from the student model and is used to generate pseudo labels. The student model receives both pseudo labels and labeled data for training. The teacher-student architecture implements stages of pre-training, pseudo annotation generation, and mixed training to achieve detection over large areas with minimal data. However, the quality errors in pseudo labels generated by the pseudo annotation generation stage accumulate with increasing training iterations, leading to irreversible adverse outcomes. This is known as the phenomenon of confirmation bias.

To enhance the quality of pseudo-labels and address the resultant confirmation bias, numerous methods have been developed, primarily categorized into framework optimization methods and soft labeling methods. Framework optimization methods refer to those that aim to optimize the structure of semi-supervised models or introduce new rules, such as Mean-teacher \cite{tarvainen2017mean}, Unbias-teacher \cite{liu2021unbiased,liu2022unbiased}, and Soft-teacher \cite{xu2021end}. Among them, Mean-teacher introduces the EMA (Exponential Moving Average) rule. Unlike traditional teacher-student architectures that share training weights, in the Mean-teacher framework, the weights of the student model are generated through exponential moving average to form the teacher model, making the overall framework more lightweight and enhancing the performance of models. Building on EMA, Unbiasd-teacher, through a two-stage framework, uses pseudo labels to train the RPN network, alleviating overfitting caused by cognitive biases to some extent; Soft-teacher proposes a more flexible strategy that resolves the issue of positive sample loss due to high thresholds, further improving model robustness. Soft labeling methods, as opposed to directly predicting pseudo-labels on images, lean towards using smoother prediction rules. For instance, Humble-teacher \cite{tang2021humble} proposes a semi-supervised framework based on soft pseudo labels, obtaining labels from the class probabilities and offset prediction distributions of detection boxes in the training process. Dense-teacher \cite{zhou2022dense} introduces depth estimation methods as an alternative to traditional pseudo-label generation methods, using depth prediction results to form constraints and enhance the quality of pseudo labels.

In summary, numerous methods have been developed to address the issue of cognitive bias in approaches based on pseudo-labeling. However, these methods still rely too heavily on labels predicted by the model, and for relatively challenging tasks such as building function recognition, pseud labels become even less reliable. Therefore, semi-supervised methods in specific scenarios still have significant room for improvement. Moreover, semi-supervised methods are typically applied within the same categorization system \cite{jiang2022prediction,yu2022pixel,shu2022mtcnet,li2023semi} and lack cross-categorization system studies, indicating substantial potential for enhancing the generalization capabilities of semi-supervised models.

\section{Datasets}

\begin{figure*}[ht]
    \centering
    \includegraphics[width=0.9\linewidth]{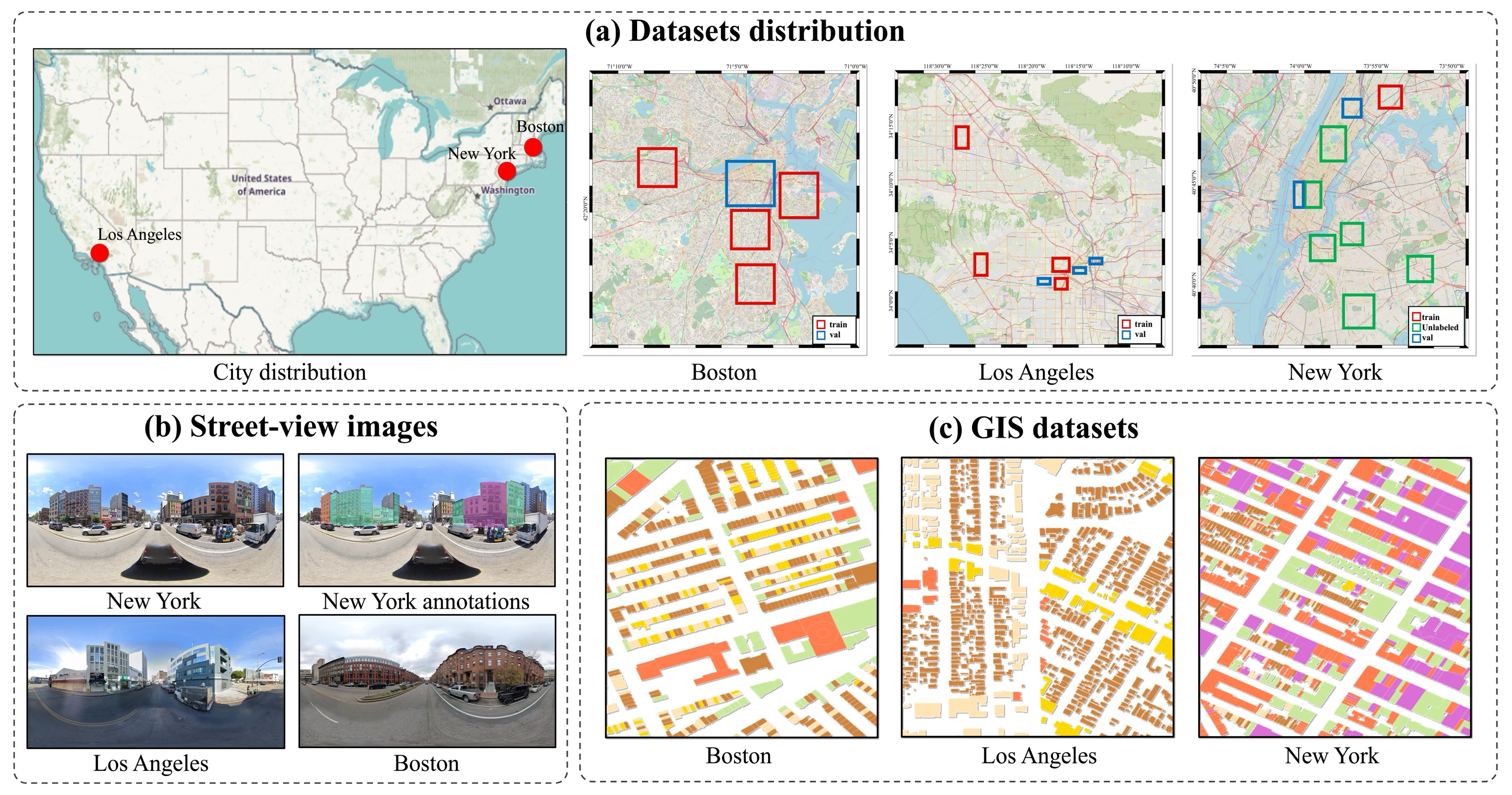}
    \caption{The overview of the datasets used in this paper. (a) The data for this study primarily come from three cities in the United States, including New York (OmniCity \cite{li2023omnicity}), Los Angeles, and Boston. (b) Street-view images from the New York all include pixel-level annotations of fine-grained building functions, while the Los Angeles and Boston areas are entirely new regions with only a few annotations available. (c) GIS data consist of building function information and footprint information, where the attribute information comes from datasets publicly available from various governments, and the building footprint information is sourced from OSM.}
    \label{fig:data}
\end{figure*}

\begin{table*}[!t]
    \centering
    \caption{Building function details of New York, Los Angeles, and Boston. The classification systems are referenced to the public datasets provided by the governments of each city.}
    \begin{tabular}{cccc}
         \toprule
         Categories & New York & Los Angeles & Boston \\
         \midrule
         1 & 1/2 family Buildings (C1)& Industrial \& Manufacturing (L1) & Single Residential (B1) \\
         2 & Walk-up Buildings (C2)& Mixed Residential/Commercial (L2) & Multiple Residential (B2) \\
         3 & Elevator Buildings (C3)& Low Medium Residential (L3) & High Residential (B3) \\
         4 & Mixed-up Buildings (C4)& Low Residential (L4) & Commercial (B4) \\
         5 & Office Buildings (C5)& Others (L5) & Others (B5) \\
         6 & Industry/Transportation/Government Facilities (C6)& - & - \\
         \bottomrule
    \end{tabular}
    \label{tab: category}
\end{table*}
\subsection{Street-view dataset}\label{svi}

There are currently many public street-view datasets available, which are used in various aspects of urban tasks \cite{tang2019cityflow,wang2016torontocity,fan2022multilevel}. In our previous work, we introduced the OmniCity dataset \cite{li2023omnicity}, designed for omnipotent city understanding with multi-level and multi-view images. This dataset comprises over 100K accurately annotated images (in COCO format), covering 25K geographical locations in New York City. It consists of three sub-datasets: a street-level panoramic dataset, a mono-view dataset (sourced from Google Street View images), and a satellite-level imagery dataset (sourced from Google's high-resolution satellite data), all geographically correlated with each other. The OmniCity dataset is now publicly available and can be found at \url{http://city-super.github.io/omnicity/}.

This study primarily focuses on the panoramic street-view dataset from the OmniCity dataset. We have adjusted and supplemented the dataset according to the task requirements, ultimately comprising a labeled training set with 3,100 images for semi-supervised training, a validation set with 3,300 images, and an Unlabeled set with 65,974 images. We have also collected street-view images from two new cities in the United States, Los Angeles and Boston, for conducting research on cross-categorization system fine-grained building function recognition. Each panoramic street-view image used in this work is \(1024\times2048\) in size and contains information on geographic coordinates, image acquisition time, panorama ID, north rotation, and zoom level. The details are shown in Figure \ref{fig:data} (a) and (b). 

\subsection{Building function and footprint information}\label{GIS}
The function information of buildings in the New York dataset originates from the government's public dataset PLUTO\footnote{\url{https://data.cityofnewyork.us/City-Government/Primary-Land-Use-Tax-Lot-Output-PLUTO-/64uk-42ks}}. By analyzing the similarity between categories and the number of samples, we merge the sixth and seventh categories which originally classified by OmniCity dataset. The final details of the building function categories of New York are shown in Table \ref{tab: category}. The building function information for Los Angeles is sourced from the publicly available GIS dataset The General Plan Land Use\footnote{\url{https://data.lacity.org/Housing-and-Real-Estate/General-Plan-Land-Use/tks4-u9wn}} by the Los Angeles City government. Overall, 35 areas were designated for planning, with a detailed classification of function displaying a clustered spatial distribution. We analyze each category and, based on the number of categories and the similarity of features between categories, merge them into 5 classes. The building function information in Boston is derived from the government dataset Analyze Boston\footnote{\url{https://data.boston.gov/dataset/boston-buildings-with-roof-breaks}}. Initially, this dataset categorized buildings into 16 different types. After conducting similar analyses on quantity, distribution, etc., as done with Los Angeles, these are ultimately consolidated into 5 types of building functions. The details of the categories for Los Angeles and Boston are shown in Table \ref{tab: category}.

\subsection{Street-view image annotation}\label{GUI}

In our previous work \cite{li2023omnicity}, we introduced a street-view image annotation tool called CityAnnotator. This tool can simulate the shooting scene of panoramic street-view images, enabling the conversion between panoramic and mono-view images. Annotations made in the mono-view can also be transformed back to the panoramic view, thus achieving high-precision annotation of panoramic images even in the presence of distortion. In addition, CityAnnotator also supports annotating three-dimensional layouts from two-dimensional scenes. 

In this work, to validate the capability of our proposed geometry-aware semi-supervised framework in fine-grained building function recognition across different categorization systems, we use this tool to construct labeled datasets for two new cities, Los Angeles and Boston. During the construction process, we standardize the annotation criteria, focusing on the building facades at the center of the street view and ignoring buildings with significant deformation or severe occlusion at the edges. Additionally, we performed a filtering process during annotation, removing street-view images with few effective building facades, ultimately creating a tiny high-quality dataset. The source code of CityAnnotator will be publicly released.

\section{Methods}

\subsection{Overall framework}

In this section, we will first provide an overview of our geometry-aware semi-supervised framework for fine-grained building function recognition. As illustrated in Figure \ref{fig:pipeline}, the geometry-aware semi-supervised framework consists of three consecutive stages, the first stage of online pre-training, the second stage of offline label generation, and the third stage of online fine-grained building function recognition. Specifically, in the first stage, we employ a end-to-end semi-supervised framework and set up a completely online pre-training process. In the second stage, we design an offline coarse annotation generation module. Based on the angular relationship between existing GIS data and panoramic street-view images, we map the building functional information from GIS data to the street-view perspective. This, combined with the building facade information (bounding boxes) obtained after decoupling the prediction results, ultimately generate coarse annotations. In the third stage, these coarse annotations and labeled data are fed into a one-stage object detection network to achieve fine-grained building function recognition. The details of each stage are introduced as follows.

\subsection{Online pre-training for building facade recognition}

\subsubsection{Semi-supervised network architecture}

Semi-supervised models based on the teacher-student architecture often set a warm-up stage at the beginning of training, during which only the student model participates in the backward update. The purpose is to first allow the model to learn from a small amount of labeled data, thereby generating a more reliable teacher model. In this work, we separate the warm-up stage, forming it into an independent pre-training stage. Compared to the usual warm-up stage, which is fully-supervised training, we opt for semi-supervised training, training both labeled and unlabeled data. The advantage of this approach lies not only in expanding the scale of the task during the warm-up stage but also in learning building-related features from unknown areas in advance. Additionally, it's worth noting that this stage remains a completely online process.

In this stage, both the teacher and student models are set to have the same structure, using RetinaNet \cite{lin2017focal} as the detection network, and leveraging a ResNet-101 backbone network pre-trained on ImageNet \cite{deng2009imagenet} to extract feature maps. Unlabeled data, before being input into the teacher model for pseudo annotation generation, first undergoes weak augmentation. After obtaining the prediction results, inspired by Consistent-teacher \cite{wang2023consistent}, we use a probability rule to filter out pseudo labels that are friendly to the student model. Subsequently, the unlabeled data goes through another round of strong augmentation and, along with the pseudo labels, is input into the student model until the model converges. 

\subsubsection{Semi-supervised network training}

The training loss in the first stage can be viewed as consisting of a loss with labeled data and a loss with unlabeled data, as shown in the Eq. \ref{eq:first_loss}. 
\begin{equation}\label{eq:first_loss}
    L_{f_{stg}}=\alpha L_l^{f_{stg}}+\beta L_u^{f_{stg}}
\end{equation}
Where \(L_l^{f_{stg}}\) and \(L_u^{f_{stg}}\) denote the loss of labeled data and the loss of unlabeled data in the first stage, \(L_{f_{stg}}\) denotes the total loss of the first stage. While \(\alpha\) and \(\beta\) represent the weights of each part of the loss. Now given labeled street-level images \(D_L={\{x_i^l,y_i^l\}}^N\) and unlabeled SVIs \(D_U={\{x_j^u\}}^M\), each training batch comprises data from both datasets according to a pre-determied ratio. After data augmentation, each batch undergoes a feature extraction layer, obtaining a feature map. The two parts of the above loss can be denoted as:
\begin{equation}
    L_l^{f_{stg}}= \sum_{i=1}^N{[L_{cls}^l(x_i^l,y_i^l)+L_{reg}^l(x_i^l,y_i^l)]}
\end{equation}
\begin{equation}
    L_u^{f_{stg}}= \sum_{i=1}^M{[L_{cls}^u(x_j^u,y_j^p)+L_{reg}^u(x_j^u,y_j^p)]}
\end{equation}
where \(x_i^l\) and \(y_i^l\) are sampled from \(D_L\), \(x_j^u\) is sampled from \(D_U\), \(y_j^p\) is a pseudo-label obtained by inference from the teacher's model, and the loss function of the two parts consists of two sub-loss functions, classification loss (\(L_{cls}\)) and regression loss (\(L_{reg}\)).

\subsection{Offline building function annotation generation}

In standard semi-supervised frameworks, the teacher model generated by the student model using EMA \cite{tarvainen2017mean} begins to participate in training after the warm-up stage, directly acting on unlabeled data to generate pseudo labels. However, the quality of pseudo labels predicted directly by the model can accumulate errors during the training process, ultimately having a significant impact on the final model. Therefore, to enhance the quality of pseudo annotations more effectively, this study introduces a coarse annotation generation module. It integrates more precise semantic information from GIS data, replacing the previously unstable category prediction results. Subsequently, by combining the semantic information with the decoupled facade information of buildings based the angular relationship between cross-view data, we generate coarse annotations of higher quality. This module specifically includes two parts, i.e., cross-view data transformation, and cross-view data matching and filtering.

\subsubsection{Cross-view data transformation}

GIS data can provide a wealth of semantic information about buildings, detailed down to each individual structure. Through GIS data, we can obtain not only the fine-grained function of buildings but also the topological relationships between features from a top-down perspective. The first stage of the coarse annotation generation module proposed in this study will fully utilize the angular relationship between GIS data and street-view images, achieving the transformation of building semantic information in GIS data to the street-view perspective.

\begin{figure}
    \centering
    \includegraphics[width=\linewidth]{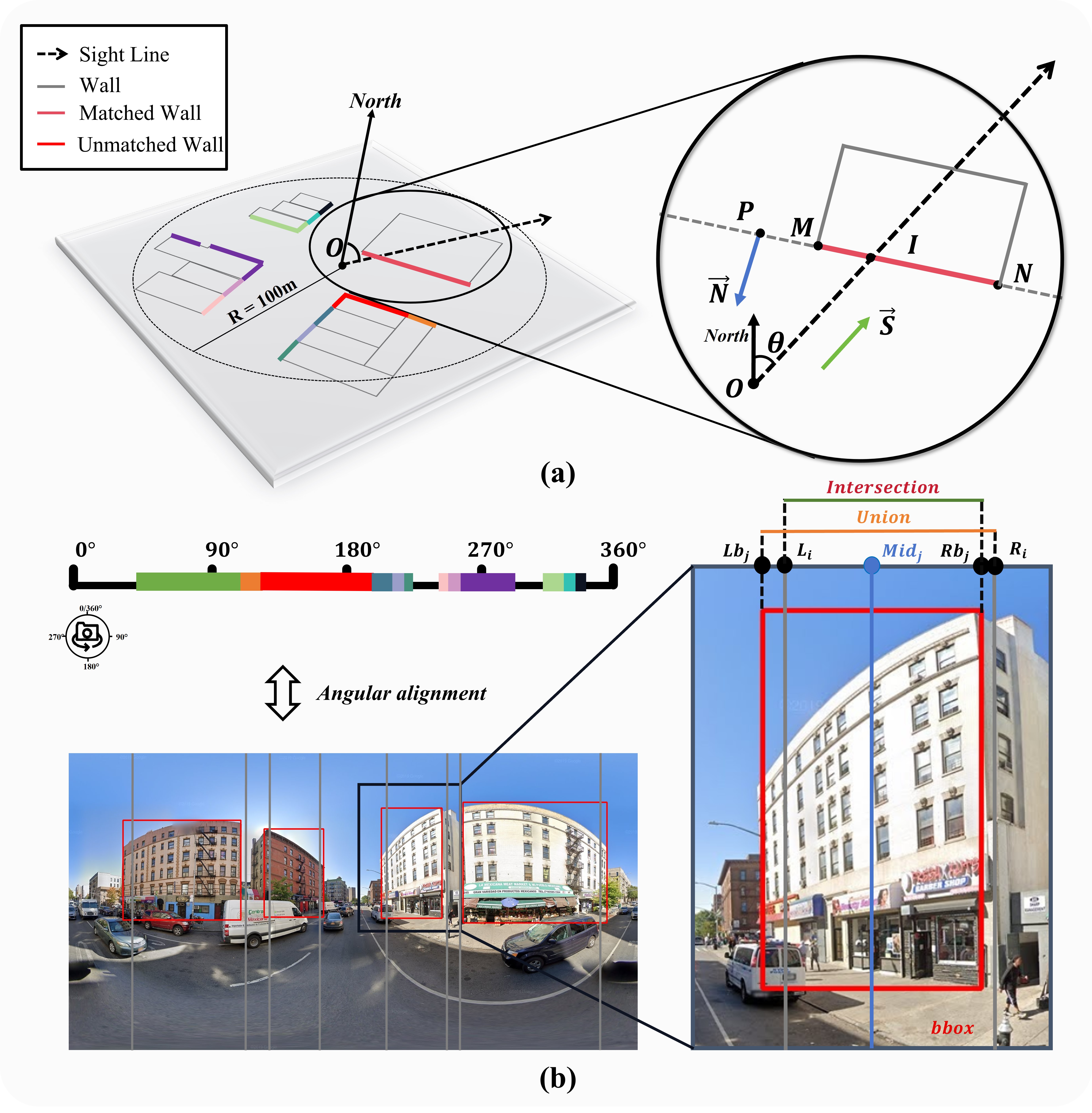}
    \caption{Coarse annotation generation module. (a) Fine-grained building function extraction from GIS data based on ray-tracing method, (b) Coarse annotation generation and filtering based on angular relationships.}
    \label{fig:cag}
\end{figure}

In practice, a panoramic camera cannot capture objects at an infinite distance, it has a set range of visibility. Our study primarily focuses on the buildings that fall within this range. We first set the position of the street-view camera, denoted as \(O\), based on the longitude and latitude of the street-view image capture point. Then, we quantify the aforementioned field of view into a circular area with \(O\) as the origin and a radius \(R\). Lastly, we reflect this onto the data space of GIS data, as shown in Figure \ref{fig:cag} (a). For the radius \(R\), we converted the longitude and latitude distances using the Eq. \ref{eq.transform}.
\begin{equation}
    \left\{
    \begin{array}{ll}
        \Delta lot \times \cos(lat) = \frac{180 \times \Delta X}{\pi \times R_E} \times 10^{-3} \\
        \Delta lat = \frac{180 \times \Delta Y}{\pi \times R_E} \times 10^{-3}
    \end{array}
    \right.
    \label{eq.transform}
\end{equation}
Where \(\Delta X\) and \(\Delta Y\) represent the Euclidean distances in GIS map view, \(\Delta lot\) and \(\Delta lat\) represent the differences in longitude and latitude, respectively, between the target point and the street-view capture point. \(R_E\) denotes the radius of the Earth, taking \(6,371.393 km\).

Subsequently, our study primarily employs a ray tracing algorithm \cite{suffern2016ray} in GIS's two-dimensional plane to match each angle within the camera's field of view with a unique building. In GIS data, all geographical entities are converted into two-dimensional vector data. Assuming the wall of a building facing the street is denoted as \(W\), with infinite length, and \(P\) represents a certain point on the wall, with \(M\) and \(N\) respectively indicating the two endpoints of a facade. At this time, a ray emitted from point \(O\) simulates the camera's line-of-sight, intersecting the wall at point \(I\). The distance set \(d\) between the wall and the shooting point under this line-of-sight can be determined using the principle of similar triangles:
\begin{equation}
    \textbf{d}=\frac{(\vec{OP}\cdot\vec{N})}{\vec{S}\cdot\vec{N}}\times\left|\vec{S}\right|
    \label{eq:d}
\end{equation}
Where \(\vec{N}\) denotes the unit normal vector of the wall, \(\vec{S}\) denotes the unit vector of the line-of-sight direction (\(\vec{OI}\)), and \(\vec{OP}\cdot\vec{N}\) denotes the projection in the normal vector direction \(\vec{OP}\), and similarly \(\vec{S}\cdot\vec{N}\) denotes the projection of the line-of-sight unit vector in the normal vector direction. Taking into account the actual situation, we select the smallest value of \(d\) from the set to determine the closest matched wall, and thereby uniquely identify the street-facing building that matches the line-of-sight.

Finally, we set the step size to \(1^\circ\) and carried out a \(360^\circ\) traversal of the line-of-sight, obtaining 360 ray tracing results. These elements are laid out in a one-dimensional coordinate system, as shown in Fig \ref{fig:cag} (b), where different building walls intersecting with the line-of-sight are intuitively represented (in different colors). Then, using a known angular information, we can match the ray tracing results with the target buildings in the street-view image, namely the north pointing angle (\(North\)). The north pointing angle is a pixel position on the image's horizontal axis, which marks a specific direction of the panoramic image when it was taken, corresponding to the true north direction geographically. In summary, based on the ray tracing algorithm, we can analyze the visibility of buildings within the camera's field of view and implement the perspective transformation of GIS data based on angular information, thus obtaining detailed function information of buildings from the street-view perspective.

\subsubsection{Cross-view data matching and filtering}\label{sec: hybrid}

After multi-classes pre-training, the model has become adept at distinguishing the facade features of buildings under different classes, and achieving high precision in target localization. However, there remains significant uncertainty in the prediction results of the buildings' functions due to the complicated features.
Therefore, in the second stage of this module, we will decouple the results predicted by the pre-trained model, separating the bounding boxes and predicted categories, and focus solely on the higher-precision building facade information (Bounding box). Subsequently, by assigning the building function information converted from GIS data in the first stage and applying multiple filtering rules, we ultimately obtain high accuracy coarse annotations.

Firstly, for the decoupled bboxes, we noticed that selecting a fixed threshold leads to the loss of a significant amount of useful information. To mitigate this loss, this study designs a step for adaptive obtaining dynamic thresholds based on probabilistic rules. We randomly divide the input street-view images into several batches in a certain quantity. A probability model is fitted using the previous batch, and then the obtained model parameters are converted into thresholds for the next batch, which are applied as constraints in the subsequent batch.

After collecting the facade information of buildings, we need to achieve more refined matching with fine-grained building function information. To this end, we have designed two matching rules:
\begin{equation}
    \left\{
    \begin{array}{cc}
    L_i<Mid_j<R_i \\
    IoU_x=\frac{Intersection}{Union}>0.3
    \end{array}
    \right.
\end{equation}
In which \(L_i \sim R_i\) represent the angular interval of a building's wall in the ray tracing results, and \(Mid_j\) represents the midpoint position of the detection box. 
\(Intersection\) and \(Union\) respectively denote the intersection and union between the ray tracing results and the detection box. Since the final result of the ray tracing lacks upper and lower bounds, the intersection over union ratio here, denoted as \(IoU_x\), is actually the ratio of intersection to union for one-dimensional line segments. By these two rules, we assume that when the midpoint of a bbox is located within the angular interval of a building's wall as per the ray tracing results, and at the same time, the one-dimensional intersection over union ratio exceeds a certain threshold, then the facade information and function information of the building can be correctly matched.

It is noteworthy that the decoupling step in the second stage of the coarse annotation generation module focuses the framework solely on the common positional information of all buildings. This has completely separated it from the attribute space of the buildings. By combining with the building function information from different regions' GIS data converted in the first stage, it facilitates the accomplishment of cross-categorization systems tasks.

\begin{table*}[!t]
    \centering
    \caption{Quantitative comparison between our method and fully-supervised methods on the New York dataset with a 1:10 labeled-unlabeled ratio. The upper part for one-stage object detection models, and the lower part for two-stage object detection models.}
    \begin{tabular}{ccccccccccccc}
    % {@{}c@{\hspace{8pt}}c@{\hspace{2pt}}c@{\hspace{2pt}}c@{\hspace{2pt}}c@{\hspace{2pt}}c@{\hspace{2pt}}c@{\hspace{9pt}}c@{\hspace{9pt}}c@{\hspace{9pt}}c@{\hspace{9pt}}c@{\hspace{9pt}}c@{\hspace{9pt}}c@{}}
        \toprule
        \multirow{2}{*}{Methods} & \multicolumn{6}{c}{Overall metrics (\%)} & \multicolumn{6}{c}{AP of each categories (\%)} \\
        \cmidrule(r){2-7} \cmidrule(r){8-13}
        & \(mAP\) & \(mAP_{50}\) & \(mAP_{75}\) & \(mAP_s\) & \(mAP_m\) & \(mAP_l\) & C1 & C2 & C3 & C4 & C5 & C6 \\
        \midrule
        RetinaNet \cite{lin2017focal} & 17.1 &28.2 &17.9 &5.0 &31.5 &19.2 &8.4 &17.3 &17.5 &26.0 &18.9 &14.5 \\
        FCOS \cite{Tian_2019_ICCV} & 18.9 & 28.8 & 20.4 & 7.1 & 33.6 & 20.2 & 11.6 & 20.6 & 18.1 & 28.6 & 19.0 & 15.8 \\
        GFL \cite{li2020generalized} & 19.8 & 28.5 &21.1 &7.5 &40.1 &20.8 &10.8 &22.7 &21.2 &30.2 &17.7 &16.1 \\
        \midrule
        Faster R-CNN \cite{ren2015faster} & 17.8 & 28.2 & 19.2 & 1.0 & 29.3 & 18.8 & 9.4 & 21.7 & 17.4 & 26.9 & 15.7 & 15.4 \\
        Cascade R-CNN \cite{cai2018cascade} & 17.7 & 26.3 & 19.0 & 1.5 & 30.3 & 18.6 & 7.7 & 22.0 & 17.4 & 27.0 & 16.6 & 15.2 \\
        Dynamic R-CNN \cite{zhang2020dynamic} & 17.1 & 26.1 & 18.8 & 0.4 & 27.7 & 18.2 & 6.5 & 21.4 & 16.8 & 26.0 & 16.1 & 15.5 \\
        \midrule
        Ours & \textbf{27.4} &\textbf{36.6} &\textbf{29.6} & \textbf{8.6} &\textbf{46.9} &\textbf{29.8} &\textbf{23.7} &\textbf{36.4} &\textbf{22.8} &\textbf{35.0} & \textbf{24.1} &\textbf{22.5} \\
        \bottomrule
    \end{tabular}
    \label{tab:result-fs}
\end{table*}
\begin{figure*}[!t]
    \centering
    \includegraphics[width=0.9\linewidth]{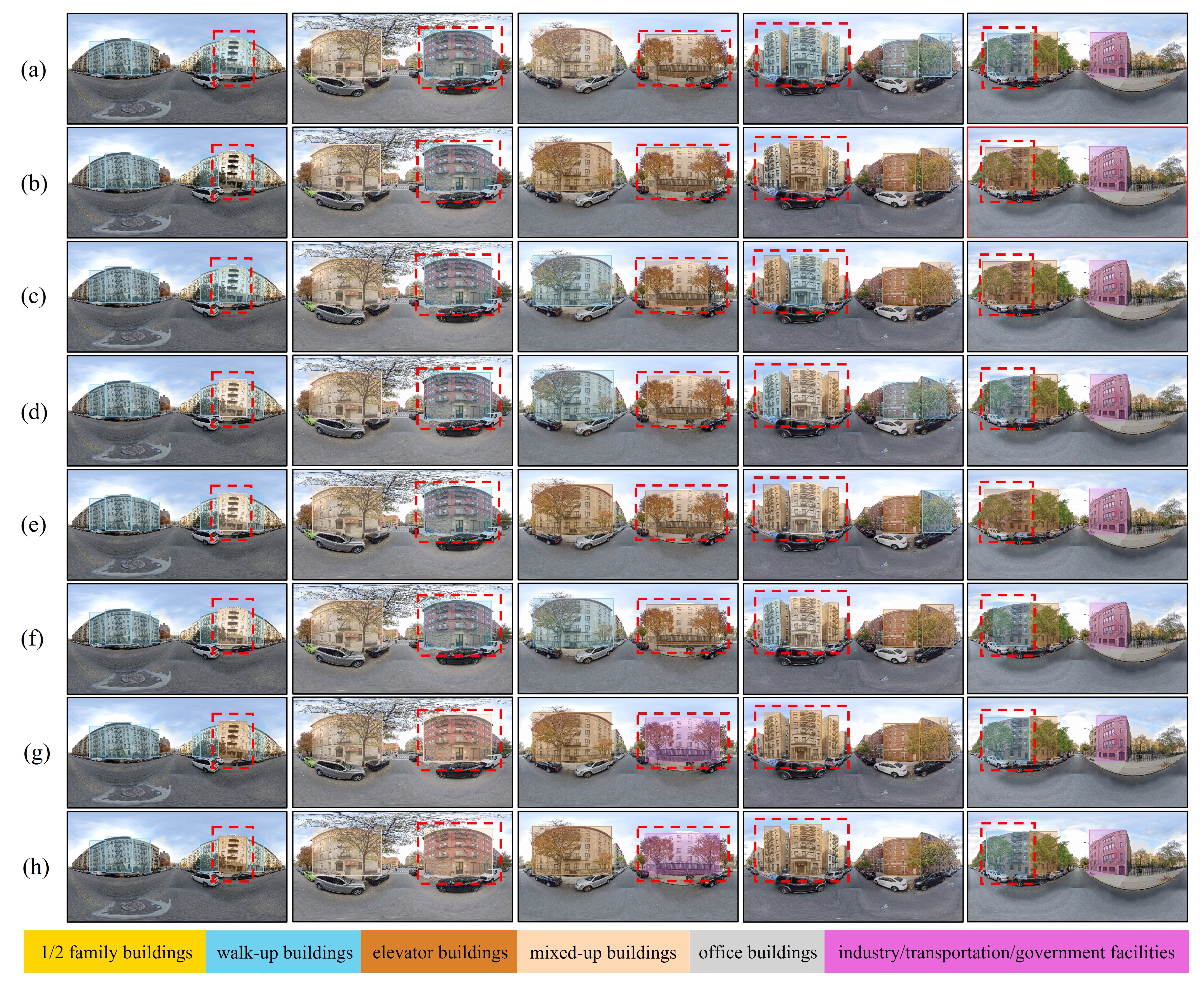}
    \caption{Visual comparisons of our method and fully-supervised methods on the fine-grained building function recognition task with the same labeled data. (a) - (c) are respectively the visual results of RetinaNet, FCOS, and GFL, which utilize the one-stage detection strategy. (d) - (f) are respectively the visual results of Faster R-CNN, Cascade R-CNN, and Dynamic R-CNN, which apply the two-stage detection strategy. (g) and (h) represent the results of our method and the ground truth, respectively.}
    \label{fig:result-rs}
\end{figure*}
\begin{table*}[!t]
    \centering
    \caption{Quantitative comparison between our method and state-of-the-art semi-supervised methods on the New York dataset with a 1:10 labeled-unlabeled ratio.}
    \begin{tabular}{ccccccccccccc}
    % {@{}c@{\hspace{8pt}}c@{\hspace{2pt}}c@{\hspace{2pt}}c@{\hspace{2pt}}c@{\hspace{2pt}}c@{\hspace{2pt}}c@{\hspace{9pt}}c@{\hspace{9pt}}c@{\hspace{9pt}}c@{\hspace{9pt}}c@{\hspace{9pt}}c@{\hspace{9pt}}c@{}}
        \toprule
        \multirow{2}{*}{Methods} & \multicolumn{6}{c}{Overall metrics (\%)} & \multicolumn{6}{c}{AP of each categories (\%)} \\
        \cmidrule(r){2-7} \cmidrule(r){8-13}
        & \(mAP\) & \(mAP_{50}\) & \(mAP_{75}\) & \(mAP_s\) & \(mAP_m\) & \(mAP_l\) & C1 & C2 & C3 & C4 & C5 & C6 \\
        \midrule
        Mean-teacher \cite{tarvainen2017mean} & 22.5 &31.2 &24.0 &8.4 &33.4 &23.6 &8.7 &28.3 &19.7 &33.1 &22.8 &21.1 \\
        Soft-teacher \cite{xu2021end} & 17.8 & 26.8 & 20.1 & 0.8 & 25.9 & 18.1 & 2.7 & 19.8 & 13.5 & 27.4 & 15.3 & 19.2 \\
        PseCo \cite{li2022pseco} & 20.4 &28.5 &22.6 &\textbf{11.0} &23.7 &21.2 &8.9 &27.1 &16.7 &32.1 &16.9 &21.5 \\
        ARSL \cite{liu2023ambiguity} & 12.7 &29.5 &7.6 &5.2 &22.0 &13.8 &5.3 &13.3 &12.4 &19.4 &13.7 &12.2 \\
        Consistent-teacher \cite{wang2023consistent} & 22.6 &30.3 &24.7 &3.6 &29.9 &24.7 &7.9 &28.7 &20.2 &31.4 &\textbf{25.2} &22.3 \\
        Ours & \textbf{27.4} &\textbf{36.6} &\textbf{29.6} &8.6 &\textbf{46.9} &\textbf{29.8} &\textbf{23.7} &\textbf{36.4} &\textbf{22.8} &\textbf{35.0} & 24.1 &\textbf{22.5} \\
        \bottomrule
    \end{tabular}
    \label{tab:result-ssod}
\end{table*}
\begin{figure*}[!t]
    \centering
    \includegraphics[width=0.9\linewidth]{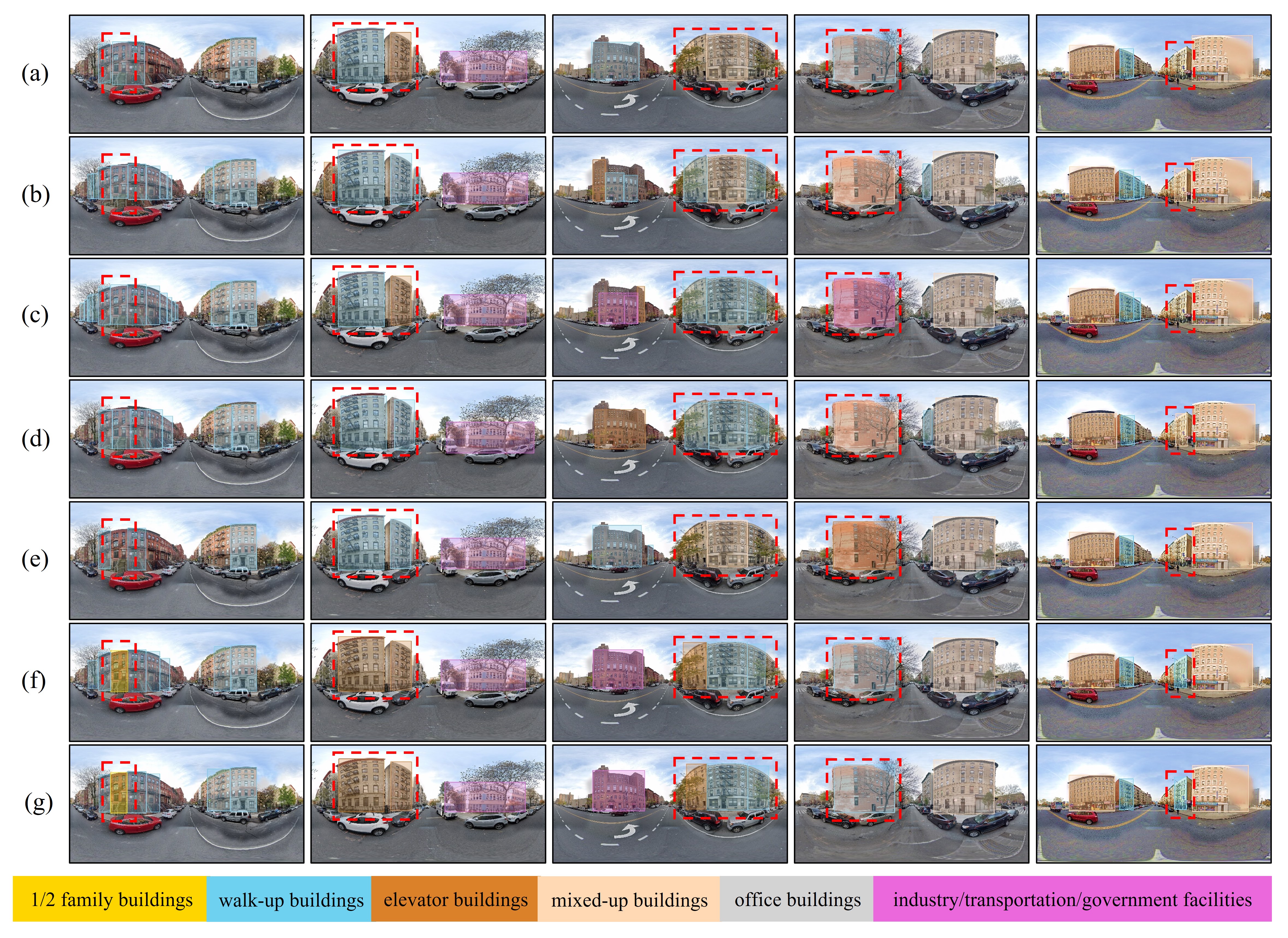}
    \caption{Visual comparisons of our method and state-of-the-art semi-supervised methods on the fine-grained building functions recognition task with $1:10$ labeled-unlabeled ratio. (a) - (e) are respectively the visual results of Mean-teacher, Soft-teacher, PseCo, ARSL, and Consistent-teacher, while (f) depicts the results of our method. (g) represents the ground truth.}
    \label{fig:result-ssod}
\end{figure*}
\subsection{Online building function recognition}

In the third stage, since the coarse annotations have already been obtained offline in the second stage, we fix the teacher model and use the loss function to only update an online one-stage object detection framework (the student model), simultaneously training both unlabeled and labeled data. 
For unlabeled data, we directly use coarse annotations to calculate the loss, similar to the calculation of loss with general pseudo labels. However, we assume that the teacher model has already converged, thus eliminating the need to generate the teacher model through EMA. Considering the imbalance of positive and negative samples in coarse annotations, we choose the Generalized Focal Loss (GFL) \cite{li2020generalized}, an improvement on Focal Loss \cite{lin2017focal}, for our loss function. GFL takes into account both the issue of sample imbalance and the quality of localization, which is very friendly for training in semi-supervised, unlabeled data regions. The loss for the third stage of training is as follows:
\begin{equation}
    L_{t_{stg}} = \alpha L_l^{t_{stg}} + \beta L_c^{t_{stg}}
    \label{stage3totalloss}
\end{equation}
\begin{equation}
    L_c^{t_{stg}} = \sum_{i=1}^{M}{[L_{cls}^c(x_j^u,y_j^c)+L_{reg}^c(x_j^u,y_j^c)]}
    \label{stage3coarseloss}
\end{equation}
where \(L_l^{t_{stg}}\) and \(L_c^{t_{stg}}\) denote the loss of labeled data and the loss of unlabeled data along with coarse annotations, respectively. \(\alpha\) and \(\beta\) represent the weights of each part of the loss. \(L_c^{t_{stg}}\) also includes both classification and regression losses. Additionally, this stage is essentially a completely online process. Through the training in this stage, we ultimately achieved large-scale, high-precision fine-grained building function recognition. It is worth noting that this stage is essentially a fully online process.

In summary, this study deconstructs the traditional semi-supervised framework into three distinct stages: online pre-training for building facade detection, building function coarse annotation generation, and building function recognition. Based on these three stages, we have achieved the integration of information from cross-view data, expanded the scope of the research, and enhanced the performance of the semi-supervised model based on pseudo annotations.

\section{Experiment}

\subsection{Experiment setting}\label{experiment}
To ensure the fairness of our experiments, we conducted all experiments on 8 NVIDIA A100 (80G) GPUs. Our proposed framework is developed based on MMDetection \cite{chen2019mmdetection}, employing a batch size of 5. The optimizer used is SGD with a learning rate of 0.01, momentum set to 0.9, and weight decay set to \(1e-4\). The dynamic parameter for EMA used to update the teacher model was set to 0.9995. Based on the basic characteristics of our dataset, we first perform pre-training for approximately 24 epochs in the first stage of the framework. Similarly, in the third stage of training, we also train for 24 epochs. We follow the data processing and data augmentation pipeline outlined in the soft-teacher framework \cite{xu2021end}. Additionally, we initiate our model using a pre-trained ResNet-101 backbone from the ImageNet \cite{deng2009imagenet} dataset. For all other semi-supervised methods and fully-supervised methods employed in our experiments, default hyperparameters are utilized.

\subsection{Evaluation metrics}\label{metrics}

In all of our experiments, we employed a set of metrics to assess the models' performance in the task of fine-grained function extraction from buildings. Specifically, we utilized three metrics adapted from the COCO evaluation metrics \cite{lin2014microsoft}. These metrics include: Average precision (\(AP\)), average precision at 50\% IoU (\(AP_{50}\)), average precision at 75\% IoU (\(AP_{75}\)), average precision for small object (\(AP_{s}\)), average precision for medium object (\(AP_{m}\)), and average precision for large object (\(AP_{l}\)).

Furthermore, to validate the performance of our proposed coarse annotation generation module, we evaluated the accuracy of the coarse annotations. Specifically, we considered a coarse annotation to be correctly matched if the intersection over union (IoU) between the bbox and the ground truth is greater than 0.8, and the categories are correctly matched based on the geometric angle relationships. Subsequently, we calculated the proportion of these correctly matched annotations among the total coarse annotations. The calculation formula is as follows:
\begin{equation}
    Accuacy=\frac{\{{bbox}_{IoU \geq thr}^c,{label}_{correct}^c\}}{{\{{bbox}_c,{label}_c\}}^M}, thr=0.8
\end{equation}
Through this index, we can intuitively see that the quality of coarse annotations is very close to manual annotations.

\subsection{Experiment results}

\subsubsection{Compare to fully-supervised methods}

We conduct a series of comparative experiments on the New York street-view dataset, which including several fully-supervised methods. The fully-supervised methods compromise one-stage RetinaNet \cite{lin2017focal}, FCOS \cite{Tian_2019_ICCV}, and GFL \cite{li2020generalized}, as well as two-stage Faster R-CNN \cite{ren2015faster}, Cascade R-CNN \cite{cai2018cascade}, and Dynamic R-CNN \cite{zhang2020dynamic}.

In the comparative experiments of this section, both the fully-supervised methods and our geometry-aware semi-supervised framework for fine-grained building function recognition utilize the same labeled data. As shown in Table \ref{tab:result-fs}, our method has achieved significant improvements compared to all the fully-supervised methods, with at least 7.6\%, 7.8\%, and 8.5\% improvements in \(mAP\), \(mAP_{50}\), and \(mAP_{75}\), respectively. This indicates that under the condition of having a smaller amount of labeled data, the performance and task scale of fully-supervised models are limited, while semi-supervised methods show better robustness and generalization ability. Additionally, we found that one-stage object detection models perform better than two-stage models in our task setting. One-stage models also have a more lightweight framework compared to two-stage models, which is why we opted for a one-stage object detection model in the third stage of our proposed method.

From the perspective of individual categories, our method has also achieved significant improvements, with the AP for C1 and C2 exceeding 10\%. Figure \ref{fig:result-rs} visualizes the results of fully-supervised methods compared with our method, where the first three columns intuitively show that, aside from our method, all other fully-supervised methods have misidentified the categories (C3 misclassified as C4, C5 misclassified as C2, C6 misclassified as C3), while the remaining two columns demonstrate that our method has more stable recognition capabilities.

\subsubsection{Compare to semi-supervised methods}

We also compared our geometry-aware semi-supervised framework with several semi-supervised methods, including the newly proposed Consistent-teacher \cite{wang2023consistent} and ARSL \cite{liu2023ambiguity}, both of which discuss the issue of pseudo annotation localization quality. In addition, we compared our method with Mean-teacher \cite{tarvainen2017mean}, PseCo \cite{li2022pseco}, and Soft-teacher \cite{xu2021end}. Mean-teacher not only modifies the teacher model but also takes into account consistency learning, while PseCo, which also considers the consistency issue, is a two-stage semi-supervised object detection framework. Soft-teacher proposes a softer solution to the threshold problem. All these semi-supervised methods \cite{wang2023consistent,tarvainen2017mean,li2022pseco,lin2014microsoft} have been commonly used as comparison methods in semi-supervised learning.

Our method has shown improvements over the state-of-the-art semi-supervised framework, Consistent-teacher, in \(mAP\), \(mAP_{50}\), \(mAP_{75}\), \(mAP_s\), \(mAP_m\) and \(mAP_l\) by 4.8\%, 6.3\%, 4.9\%, 5.0\%, 17.0\%, and 5.1\%, respectively, as shown in Table \ref{tab:result-ssod}, demonstrating better generalization ability. In detecting targets of different sizes, PseCo, proposed in 2022, exhibited superior performance in detecting small targets. However, it still lagged behind our method by 23.2\% in medium target detection and by 6.1\% in large target detection, respectively. This indicates that our method has a more balanced performance in the task of fine-grained building function recognition.

As shown in Table \ref{tab:result-ssod}, our method has also achieved certain improvements in categories, with C1 showing a 14.8\% improvement over the best-performing method. We found that, regardless of whether it is a fully-supervised or semi-supervised method, there are certain differences in the indicators between different categories, with C2 and C4 having the best detection accuracy, and C1 and C3 being relatively poor. Although we have noticed the phenomenon of sample number imbalance and have conducted relevant data processing, there is still a gap of more than 10\%. This may be related to the complexity of the features between different categories. However, our method still possesses the best performance.

Figure \ref{fig:result-ssod} shows the visualization results of our method compared with other semi-supervised methods. It can be seen that our geometry-aware semi-supervised framework has good robustness in predicting categories under different scenarios, as well as high-quality localization. Additionally, our model demonstrates high adaptability, exhibiting strong distinguishing capabilities for easily confused categories and detection targets of varying sizes.

\begin{table*}[!t]
    \centering
    \caption{Comparison results of different models for fine-grained building function recognition across various labeled-to-unlabeled ratios. The labeled data remains fixed, and we simulate different ranges of fine-grained building function recognition by varying the amount of unlabeled data ($1:1$, $1:5$ and $1:10$). Among them, our method achieves the best performance in all three task settings, indicating that our method possesses better robustness as the scale of the task increases.}
    \begin{tabular}{@{}c@{\hspace{2pt}}c@{\hspace{2pt}}c@{\hspace{2pt}}c@{\hspace{2pt}}c@{\hspace{2pt}}c@{\hspace{2pt}}c@{\hspace{2pt}}c@{\hspace{9pt}}c@{\hspace{9pt}}c@{\hspace{9pt}}c@{\hspace{9pt}}c@{\hspace{9pt}}c@{\hspace{9pt}}c@{}}
    \toprule
    \multirow{2}{*}{Methods} & \multirow{2}{*}{Labeled-unlabeled ratio} & \multicolumn{6}{c}{Overall metrics (\%)} & \multicolumn{6}{c}{AP of each categories (\%)}\\
    \cmidrule(r){3-8} \cmidrule(r){9-14}
    & & \(mAP\) & \(mAP_{50}\) & \(mAP_{75}\) & \(mAP_s\) & \(mAP_m\) & \(mAP_l\) & C1 & C2 & C3 & C4 & C5 & C6 \\
    \midrule
    GFL \cite{li2020generalized} & Labeled only & 19.8 & 28.5 & 21.1 & 7.5 & 40.1 & 20.8 & 10.8 & 22.7 & 21.2 & 30.2 & 17.7 & 16.1\\
    \midrule
    \multirow{3}{*}{Consistent-teacher \cite{wang2023consistent}} & $1:1$ & 20.0 & 27.2 & 21.4 & 7.9 & 26.5 & 21.6 & 6.0 & 28.5 & 17.8 & 30.9 & 20.6 & 16.3\\
     & $1:5$ & 21.3 & 28.4 & 22.9 & 4.2 & 25.1 & 23.2 & 4.1 & 30.7 & 17.6 & 33.0 & 23.8 & 18.8\\
     & $1:10$ & 22.6 & 30.3 & 24.7 & 3.6 & 29.9 & 24.7 & 7.9 & 28.7 & 20.2 & 31.4 & \textbf{25.2} & 22.3\\
    \midrule
    \multirow{3}{*}{Ours}  & $1:1$ & 24.5 & 33.5 & 26.4 & 8.6 & 46.6 & 25.9 & 18.7 & 31.5 & 20.3 & 33.8 & 21.0 & 21.4\\
     & $1:5$ & 26.0 & 35.0 & 27.9 & \textbf{10.1} & 45.9 & 28.0 & 22.4 & 33.5 & 20.9 & 35.0 & 23.9 & 20.1\\
     & $1:10$ & \textbf{27.4} & \textbf{36.6} & \textbf{29.6} & 8.6 & \textbf{46.9} & \textbf{29.8} & \textbf{23.7} & \textbf{36.4} & \textbf{22.8} & \textbf{35.0} & 24.1 & \textbf{22.5} \\
    \bottomrule
    \end{tabular}
    \label{tab: ratio}
\end{table*}

\section{Discussion}

\subsection{The scalability of our method using different sizes of unlabeled data}

Expanding the scale of research is of great practical significance. To further discuss the scalability ability of our proposed multi-stage semi-supervised framework for fine-grained building funs recognition at different scales, we simulate different task scales using various label-unlabeled ratios. With the number of labeled data held constant, we use label-unlabeled ratios of $1:1$, $1:5$, and $1:10$, respectively. In addition, we also compare the scenario with only labeled data (fully-supervised), with results shown in Table \ref{tab: ratio}.

We also compare our approach with a state-of-the-art semi-supervised method, setting three different ratios alike. It is evident that, in contrast to fully-supervised training limited to a small scale, semi-supervised methods achieve varying degrees of accuracy improvement as the scale of research changes. With slight improvements over the Consistent-teacher by 0.2\%, 1.5\%, and 2.8\% in \(mAP\), \({mAP}_{50}\), and \({mAP}_{75}\), our method exhibits significant enhancements of 4.7\%, 6.2\%, and 7.6\%. This reflects that our method can maintain relatively stable performance and possesses better scalability ability when facing more unknown situations in urban city and expanding the research scale.
Additionally, from the perspective of individual categories, as the scale of the study increases, our method results in a uniform increase in $mAP$ among categories without fluctuations, while the Consistent-teacher model exhibits greater variability. For example, when the ratio shifts from $1:5$ to $1:10$, the accuracy for C1, C3, C6, and C7 increases, whereas it decreases for C2 and C4. Under the same conditions, apart from C4, whose $mAP$ remains unchanged, our method sees improvements in the other five categories. This further demonstrates the stronger robustness of our proposed multi-stage semi-supervised framework.
\begin{figure*}[!t]
    \centering
    \includegraphics[width=0.9\linewidth]{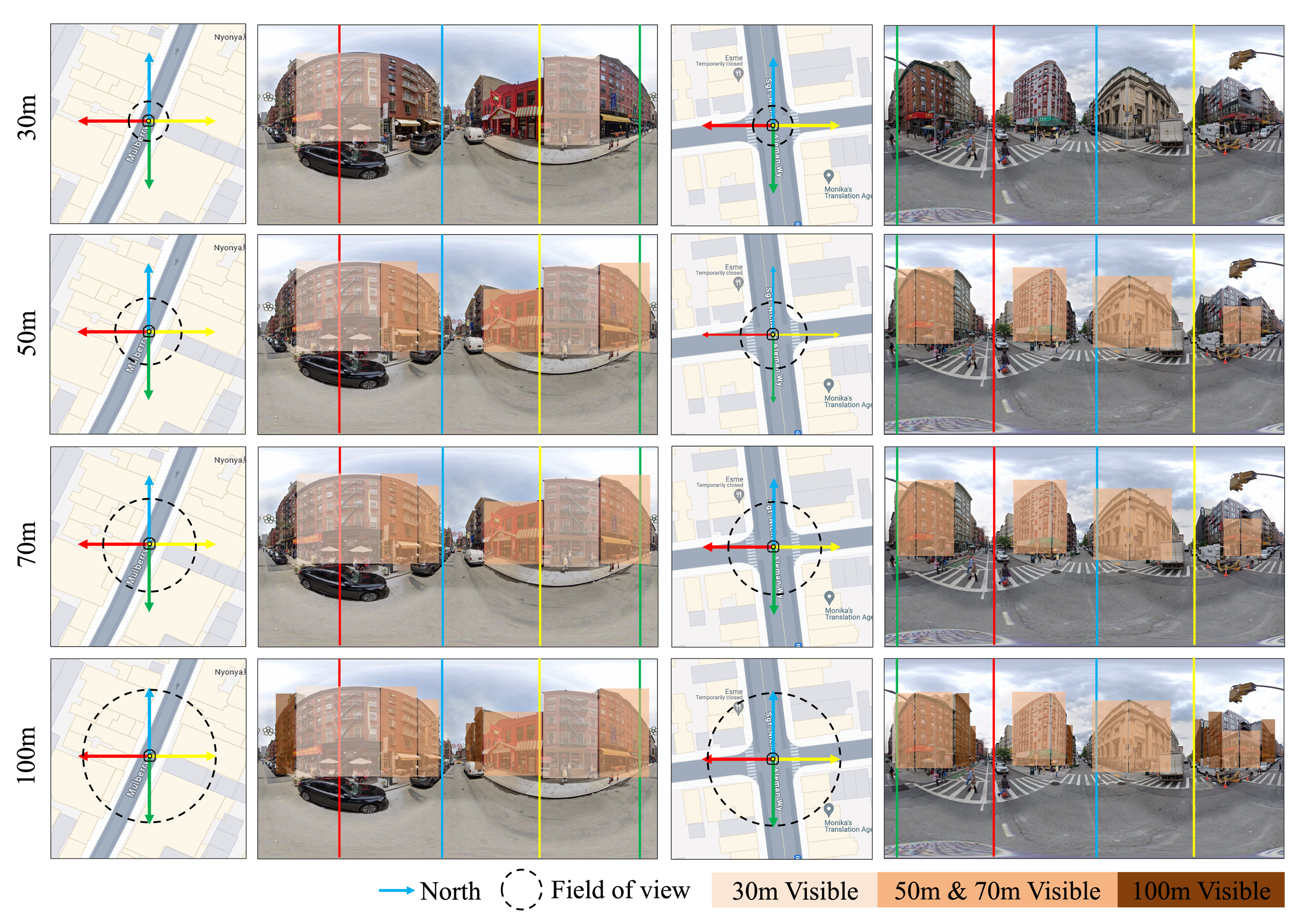}
    \caption{Visualization of building tracing results under different radius in the coarse annotation generation module. The first column of street-view images represents the situation of a single road, while the second column of street-view images represents the situation at a cross road.}
    \label{fig:region}
\end{figure*}

\subsection{Ablation study of the coarse annotation generation module}
In this section, we will analyze the important parameters of the coarse annotation generation module of our proposed multi-stage semi-supervised framework for fine-grained building functions recognition. These parameters include the field of view radius for ray tracing and the threshold for decoupling bboxes.

In the coarse annotation generation module, we quantified the visible range of a panoramic camera into a circular area with a radius \(R\) within the GIS data space. The ray tracing algorithm will primarily analyze the visibility of buildings within that range. According to the lane width standards provided in official documents, the width of a single lane in urban areas, including arterial and collector roads, is approximately 3.0-3.6 meters\footnote{\url{https://safety.fhwa.dot.gov/geometric/pubs/mitigationstrategies/chapter3/3_lanewidth.cfm}}. Assuming the roads in the city are uniformly 10m wide (double lanes and sidewalks), we conducted experiments with radius widths of 30m, 50m, 70m and 100m to study the impact of different distances on the performance of the coarse annotation generation module and the final model.

\begin{table*}[!t]
    \centering
    \footnotesize
    \caption{Ablation analysis of the coarse annotation generation module with different field of view radius ($R$) in term of COCO metrics. Among them, when $R=50$ and $R=70$, the model exhibits similar performance. However, considering the overall training efficiency of the framework, we ultimately selected $R=50m$ as the final experimental parameter.}
    \begin{tabular}{cccccccccccccc}
        \toprule
        \multirow{2}{*}{Methods} & & \multicolumn{6}{c}{Overall metrics (\%)} & \multicolumn{6}{c}{AP of each categories (\%)} \\
        \cmidrule(r){3-8} \cmidrule(r){9-14}
         & \(Accuracy\) & \(mAP\) & \(mAP_{50}\) & \(mAP_{75}\) & \(mAP_{s}\) & \(mAP_{m}\) & \(mAP_{l}\) & C1 & C2 & C3 & C4 & C5 & C6\\
        \midrule
        Ours (R=30m) & 95.6 & 21.9 & 30.0 & 23.6 & 4.0 & 36.7 & 23.8 & 2.0 & 27.4 & 16.4 & 30.5 & 18.8 & 18.2 \\ 
        Ours (R=50m) & 96.5 & \textbf{27.4} & 36.6 & \textbf{29.6} & 8.6 & 46.9 & \textbf{29.8} & \textbf{23.7} & \textbf{36.4} & 22.8 & \textbf{35.0} & 24.1 & 22.5 \\
        Ours (R=70m) & \textbf{97.0} & 27.1 & \textbf{37.2} & 29.3 & 5.7 & \textbf{47.7} & 28.9 & 20.9 & 30.4 & \textbf{24.6} & 34.8 & \textbf{27.3} & \textbf{24.6} \\
        Ours (R=100m) & 96.9 & 26.8 & 35.8 & 28.9  & \textbf{15.9} & 40.3 & 28.0 & 22.0 & 33.9 & 24.5 & 36.4 & 23.3 & 20.9\\
        \bottomrule
    \end{tabular}
    \label{tab: region}
\end{table*}
\begin{table*}[!t]
    \centering
    \footnotesize
    \caption{Ablation analysis of the coarse annotation generation module with different thresholds for filtering building facade information in terms of COCO metrics.}
    \begin{tabular}{ccccccccccccc}
        \toprule
        \multirow{2}{*}{Methods} & \multicolumn{6}{c}{Overall metrics (\%)} & \multicolumn{6}{c}{AP of each categories (\%)} \\
        \cmidrule(r){2-7} \cmidrule(r){8-13}
         & \(mAP\) & \(mAP_{50}\) & \(mAP_{75}\) & \(mAP_{s}\) & \(mAP_{m}\) & \(mAP_{l}\) & C1 & C2 & C3 & C4 & C5 & C6\\
        \midrule
        Ours (${IoU}_{thred} = 0.3$) & 23.2 & 31.5 & 24.9 & \textbf{13.5} & \textbf{49.7} & 23.8 & 20.8 & 33.8 & 17.3 & 32.4 & 17.5 & 17.5 \\
        Ours (${IoU}_{thred} = 0.5$) & 25.5 & 34.5 & 28.0 & 2.8 & 44.0 & 28.6 & 18.7 & 33.9 & 22.2 & 34.6 & 23.0 & 20.7 \\
        Ours (${IoU}_{thred} = 0.7$) & 22.0 & 30.2 & 23.9 & 0.6 & 34.8 & 25.1 & 18.3 & 28.8 & 15.7 & 31.0 & 19.1 & 19.3 \\
        Ours (self-adaptive ${IoU}_{thred}$) & \textbf{27.4} & \textbf{36.6} & \textbf{29.6} & 8.6 & 46.9 & \textbf{29.8} & \textbf{23.7} & \textbf{36.4} & \textbf{22.8} & \textbf{35.0} & \textbf{24.1} & \textbf{22.5} \\
        \bottomrule
    \end{tabular}
    \label{tab:thred}
\end{table*}

As illustrated in Figure \ref{fig:region}, when the \(R\) value is 30m, only a few buildings can be tracked in single road situation, and in the case of cross road, there are even no visible buildings. When the \(R\) value is set to 100m, it can be seen that a larger number of buildings can be tracked on the GIS map through the algorithm. However, when reflected in the street-view perspective, many of the tracked results are quite small, evidently exceeding the capture range of the panoramic street-view camera. When \(R\) is set to 50m and 70m, the tracking results of both are close to each other and are more in line with the real world.
Table \ref{tab: region} displays the impact of different \(R\) values on the metrics and the accuracy of the coarse annotations (the detail is shown in the section \ref{metrics}). It can be observed that under four different \(R\) values, the accuracy of coarse annotations both reaches over 95\%, which is very close to the ground truth. As for COCO metrics, \(R\) set at 50m and \(R\) at 70m perform better compared to the other distances, with similar levels of precision. However, considering that a larger \(R\) distance allows the ray tracing algorithm to track a wider range, choosing a relatively smaller \(R\) value of 50m will make the entire framework more efficient.

Additionally, when filtering the decoupled building facade information, we devised a strategy for an adaptive dynamic threshold adjustment based on probabilistic rules. This is because we observed that the pre-trained model's recognition results for building facades vary across different scenes, and choosing a fixed threshold could miss a lot of valuable information.

Table \ref{tab:thred} displays the comparison results of our selection between different fixed thresholds and the adaptive threshold strategy. It was found that when a fixed threshold strategy is selected, the \(mAP\) performs best at a threshold of 0.5, reaching 25.5\%, but decreases to 22.0\% at a higher threshold of 0.7. Although a fixed threshold can ensure a certain level of localization accuracy for building facades, in more complex street-view scenarios, a lot of useful information is lost due to the high threshold, ultimately leading to a decrease in model performance. Our adaptive dynamic adjustment threshold strategy, based on the probabilistic model, can enhance the scalability ability of the coarse annotation module, thereby extracting facade information from buildings in more diverse scenarios and ultimately providing the model with better robustness and scalability capability.

\subsection{Cross-regional fine-grained building functions recognition}

\setlength{\tabcolsep}{5pt}
\begin{table*}[ht]
    \centering
    \footnotesize
    \caption{Quantitative results for cross-regional fine-grained building function recognition by our geometry-aware semi-supervised method in Los Angeles and Boston comparing training. The annotations includes both data constructed by combining GIS data with a coarse annotation generation module and manually annotated data. We also conducted comparative experiments with and without the participation of manually annotations.}
    \begin{tabular}{ccccccccccccc}
         \toprule
         \multirow{2}{*}{City} & \multirow{2}{*}{Manual annotations} & \multicolumn{6}{c}{Overall metrics (\%)} & \multicolumn{5}{c}{AP of each categories (\%)} \\
         \cmidrule(r){3-8} \cmidrule(r){9-13}
          & & \(mAP\) & \(mAP_{50}\) & \(mAP_{75}\) & \(mAP_{s}\) & \(mAP_{m}\) & \(mAP_{l}\) & L1/B1 & L2/B2 & L3/B3 & L4/B4 & L5/B5 \\
         \midrule
         \multirow{2}{*}{Los Angeles} & & 31.2 & 38.6 & 33.9 & - & 17.9 & 33.5 & 33.0 & 56.0 & 46.2 & 54.7 & 3.2 \\
          & \checkmark & 33.5 & 38.6 & 33.9 & - & 19.7 & 36.4 & 44.2 & 58.6 & 46.6 & 49.4 & 4.5 \\
         \midrule
         \multirow{2}{*}{Boston} & & 32.3 & 38.6 & 34.5 & - & 11.8 & 33.8 & 25.5 & 43.0 & 32.9 & 28.6 & 31.4 \\
          & \checkmark & 32.5 & 39.0 & 34.6 & - & 13.6 & 33.9 & 26.5 & 43.4 & 33.3 & 27.1 & 32.3 \\
         \bottomrule
    \end{tabular}
    \label{tab: cross-regional}
\end{table*}
\begin{figure*}[ht]
    \centering
    \includegraphics[width=0.9\linewidth]{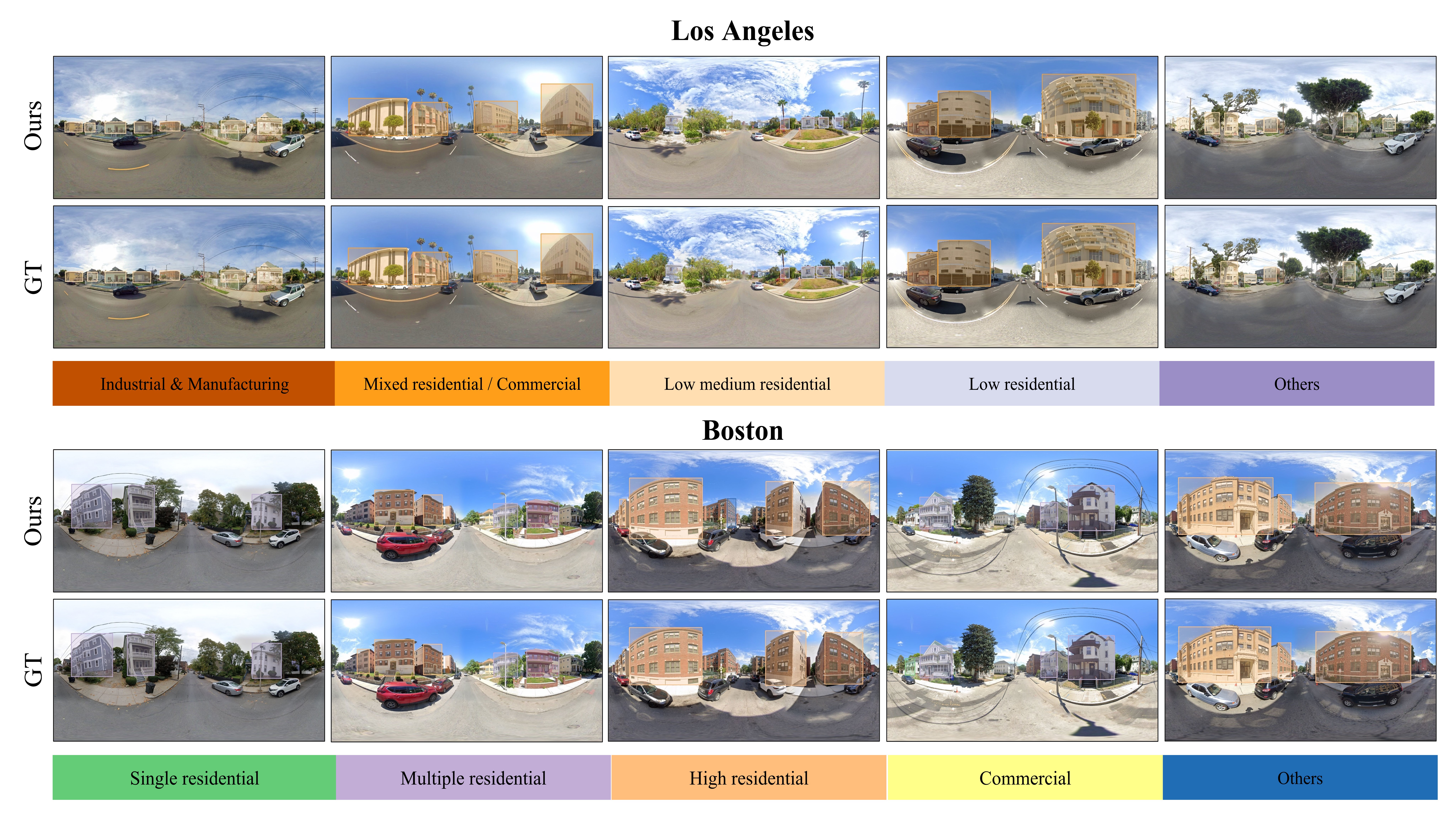}
    \caption{Visualization results of cross-categorization system fine-grained building function recognition.}
    \label{fig: cross-regional}
\end{figure*}

In this section, we will primarily explore the potential of the proposed geometry-aware semi-supervised framework for fine-grained building function recognition in cross-categorization system tasks. The main challenge of cross-categorization system tasks stems from the significantly different feature spaces across regions, including diverse facade features of buildings and different classification systems. Our multi-stage training strategy is highly flexible, adapting effectively to challenges by adjusting input data and model architecture, enabling fine-grained building function recognition across different categorization systems.

The annotated data for the new region was constructed by combining local GIS data with a coarse annotation generation module. Additionally, we simulated two scenarios for cross-categorization system tasks. In one scenario, there is some prior knowledge in new area, where a small portion of street-view images were manually annotated. In the other scenario, these two regions are entirely unknown, with no human intervention.
During the first pre-training stage, we use labeled data from New York as the labeled input and data from new regions as the unlabeled input. In the first scenario, a small amount of manual labeled data from the new regions is added to the labeled dataset, while in the second scenario, only data from New York is used. Moreover, as mentioned in the section \ref{sec: hybrid}, the multi-task pre-training model is better suited for recognizing building facades across different categories, which is insufficient for cross-categorization system tasks. Therefore, we opt for a simpler binary classification task, focusing directly on the facade information. 
In the second stage, we utilize GIS data from the target domain to extract building function information, combining them with the facade information obtained by the pre-trained model to generate coarse annotations for multi-class tasks. In the third stage, we input these coarse annotations to undertake multi-class tasks in the target domain. For the first scenario, we input both the labeled and coarse annotated data for training, while in the second scenario, only the coarse annotations are used for training.

From Table. \ref{tab: cross-regional} we can see that our framework for fine-grained building function recognition demonstrates good performance in cross-categorization system experiments. In completely unlabeled scenarios, the accuracy in the Los Angeles and Boston areas reached 31.2\% and 32.3\%, respectively, and the model also showed balanced performance across different IoU intervals. However, the framework seems less effective for detecting small and medium-sized targets. Examining individual categories, in Los Angeles, except for L5, all other \(AP\)s exceeded 30\%, with L2 and L4 even surpassing 50\%. This could be due to a lower number of L5 instances in the selected areas, leading to significant sample imbalance in the coarse annotations of L5 obtained in the second stage of the framework. In the Boston area, the performance across various categories was relatively balanced, with the lowest \(AP\) being 26.5\% for B1 and the highest being 43.4\% for B2. 
In Figure \ref{fig: cross-regional}, we display the ground truth (second row) and the visualization results of training without labeled data (first row) for two regions. The recognition and localization of fine-grained building functions are both well performed.

In summary, our framework demonstrates good robustness in cross-categorization system fine-grained building function recognition. This applies both to scenarios like Boston, where the functions distribution is relatively uniform and the building sizes are moderate, and to areas like Los Angeles, where the roads are wider and some categories of buildings are smaller.

\subsection{Comparison between human recognition and algorithm recognition}

\begin{figure*}[!t]
    \centering
    \includegraphics[width=0.9\linewidth]{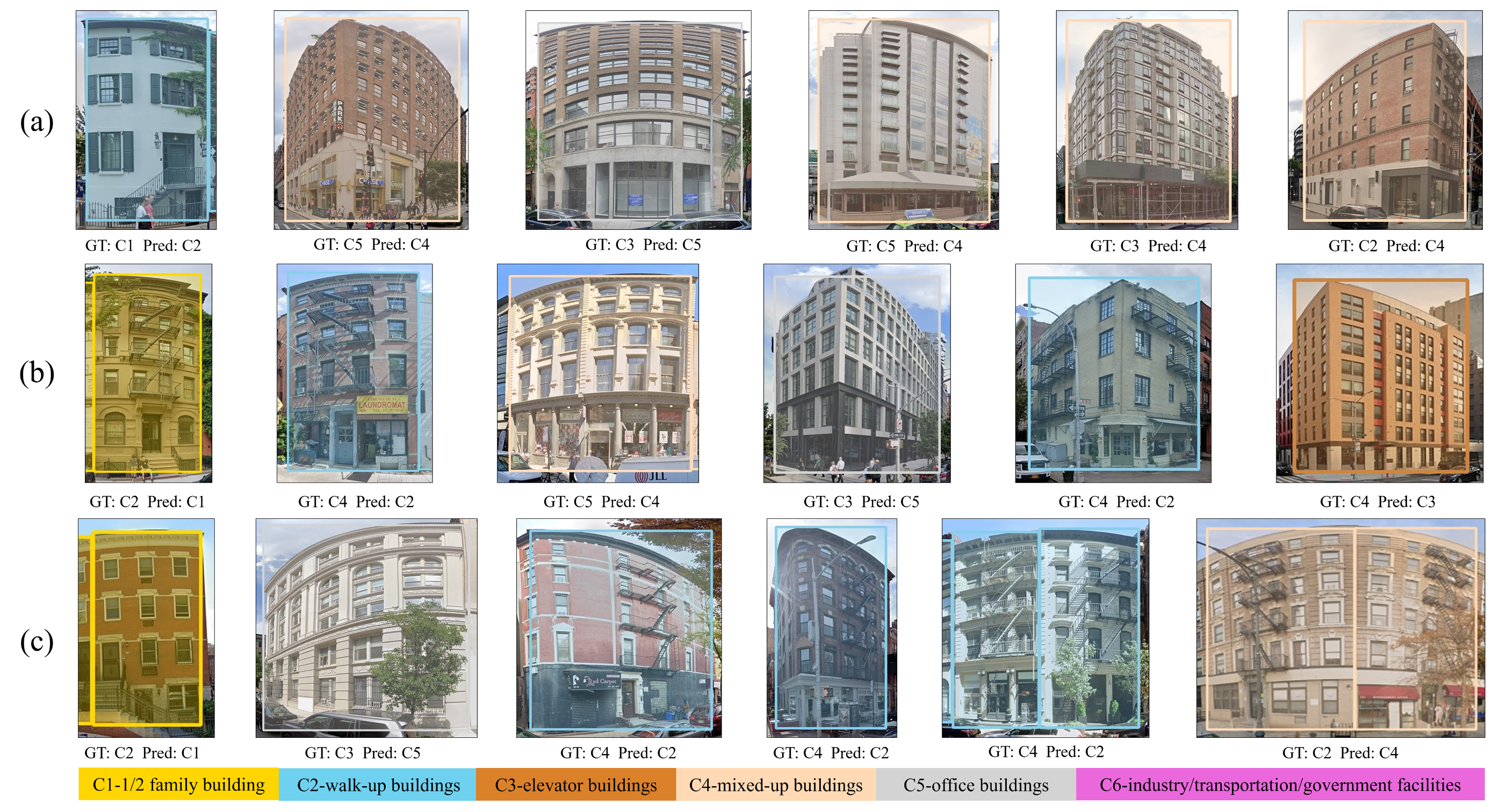}
    \caption{The hard cases for manual recognition and algorithm recognition. The row (a) lists the errors made by the algorithm recognition (our geometry-aware semi-supervised framework). The row (b) lists the errors made by human recognition. The row (c) lists the errors made by human recognition, while the algorithm recognition achieves correct results.}
    \label{fig:badcase}
\end{figure*}
\begin{figure}[ht]
    \centering
    \includegraphics[width=\linewidth]{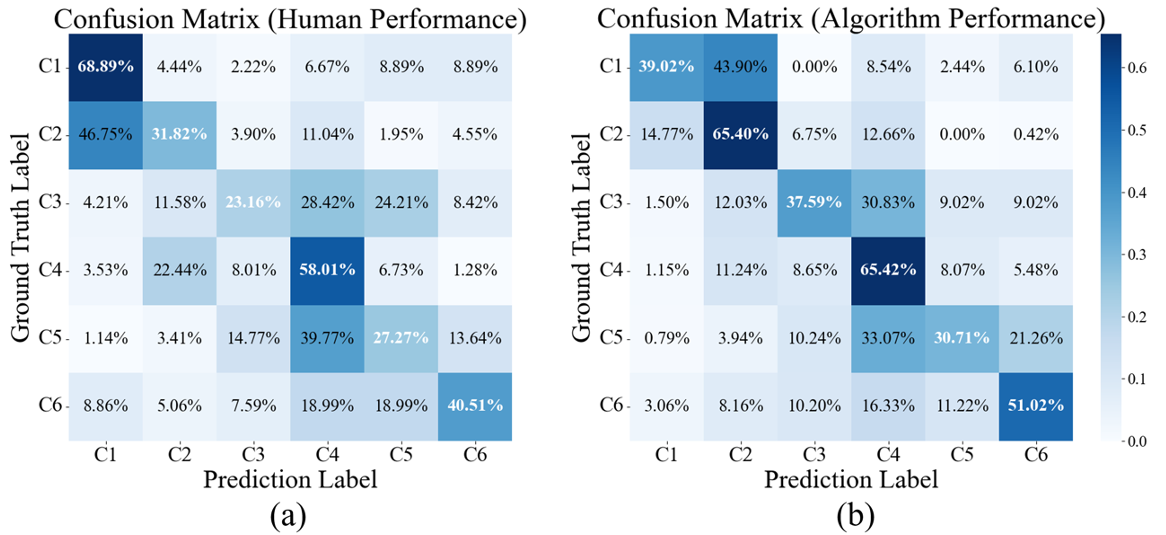}
    \caption{The normalized confusion matrices, where (a) and (b) represent the human recognition and algorithm recognition, respectively.}
    \label{fig:confusion_matrix}
\end{figure}

The experimental results indicate that our method still has certain limitations in fine-grained building function recognition task. To explore the difficulty of this task, we conduct an additional experiment involving manual recognition of building function. Several researchers are provided with training on the task of fine-grained building function recognition, learning about the categorization system of building function as well as the morphology and facade features of various types of buildings. Then the researchers are asked to perform fine-grained building function recognition on 200 images from the validation set, where the buildings were provided with bbox information only. For these 200 street-view images from the validation set, we consider not only the need to include all categories but also aim to ensure a relatively random spatial distribution. We also use the aforementioned best-performing model to make predictions on these 200 images. The resulting confusion matrices from human recognition and our method are shown in Figure \ref{fig:confusion_matrix}. 

Due to the mixed-use nature of urban functional areas during city development, most buildings have diverse functions and it is difficult to accurately distinguish certain categories. For buildings with both C1 and C2 features, which have fewer floors and staircases on the facade, the confusion between the two is common. For buildings where the first floor differs significantly from the upper floors, they are easily misrecognized as C4. Figure \ref{fig:badcase} (a) demonstrates some typical error cases. The manual recognition results have similar error cases to our method, as shown in Figure \ref{fig:badcase} (b). Conversely, our method can extract more implicit features, resulting in better accuracy than manual recognition across all categories except C1. Some cases where our method performs better than manual recognition are shown in Figure \ref{fig:badcase} (c). Additionally, our method offers significant advantages in terms of time efficiency and learning costs. Therefore, we believe that the framework proposed in this study holds great potential for fine-grained building function recognition.

\section{Conclusion}

In this study, we propose a geometry-aware semi-supervised framework for fine-grained building function recognition. This method decomposes the general teacher-student structure of semi-supervised framework into three stages. In the first pre-trained stage, we employ semi-supervised methods to learn building facade location information from a large amount of unlabeled data for precise building facade detection. In the second stage, a coarse annotation generation module based on cross-perspective data angular relationships is proposed. This module effectively utilizes existing GIS data and street-view images to generate high-accuracy coarse annotations. In the third stage, by combining the coarse annotations with other labeled data, we achieved large-scale high-precision building fine-grained function recognition under few-shot conditions, based on a object detection network.

We reorganize the OmniCity dataset for semi-supervised building fine-grained function recognition and conducted extensive experiments on this dataset. The experimental results indicate that our method achieves better building fine-grained function recognition results compared to state-of-the-art semi-supervised and fully supervised methods, with an improvement of over 5\% in \(mAP\). Additionally, we construct two new city datasets for Los Angeles and Boston using the CityAnnotator tool and discuss the performance of our method in the task of building fine-grained function recognition across different classification systems based on these datasets. We also analyze several important parameters and provided a detailed discussion on the robustness and generalization ability of the model.

In future work, we will expand the scope of our research to more cities and explore additional data types and urban elements. At the same time, we will further optimize the fine-grained building function recognition method, providing technical and data support for understanding urban morphology and analyzing human behavior.

\ifCLASSOPTIONcaptionsoff
  \newpage
\fi

\bibliographystyle{IEEEtran}
\bibliography{bibtex/bib/IEEEabrv,ref_v1}

\end{document}